\documentclass{article}

\PassOptionsToPackage{numbers, compress}{natbib}

\usepackage[preprint]{neurips_2026}

\usepackage[utf8]{inputenc}
\usepackage[T1]{fontenc}
\usepackage{hyperref}
\usepackage{url}
\usepackage{booktabs}
\usepackage{amsmath}
\usepackage{amsfonts}
\usepackage{nicefrac}
\usepackage{microtype}
\usepackage{xcolor}
\usepackage{graphicx}
\usepackage{multirow}
\usepackage{array}
\usepackage{tcolorbox}

\newcommand{\benchmark}{\textsc{FIKA-Bench}}

\title{\benchmark: From Fine-grained Recognition to Fine-Grained Knowledge Acquisition}

\author{
    Geng Li$^{1}$ \quad Yuxin Peng$^{1}$\footnotemark[2]\\
    \texttt{ligeng@stu.pku.edu.cn} \quad \texttt{pengyuxin@pku.edu.cn}\\[1mm]
    $^{1}$Wangxuan Institute of Computer Technology, Peking University, China
}

\begin{document}

\maketitle
\begingroup
\renewcommand{\thefootnote}{\fnsymbol{footnote}}
\footnotetext[2]{Corresponding author.}
\endgroup

\begin{figure*}[!h]
    \vspace{-2mm}
    \centering
    \begin{minipage}[t]{0.9\linewidth}
        \centering
        \includegraphics[width=\textwidth]{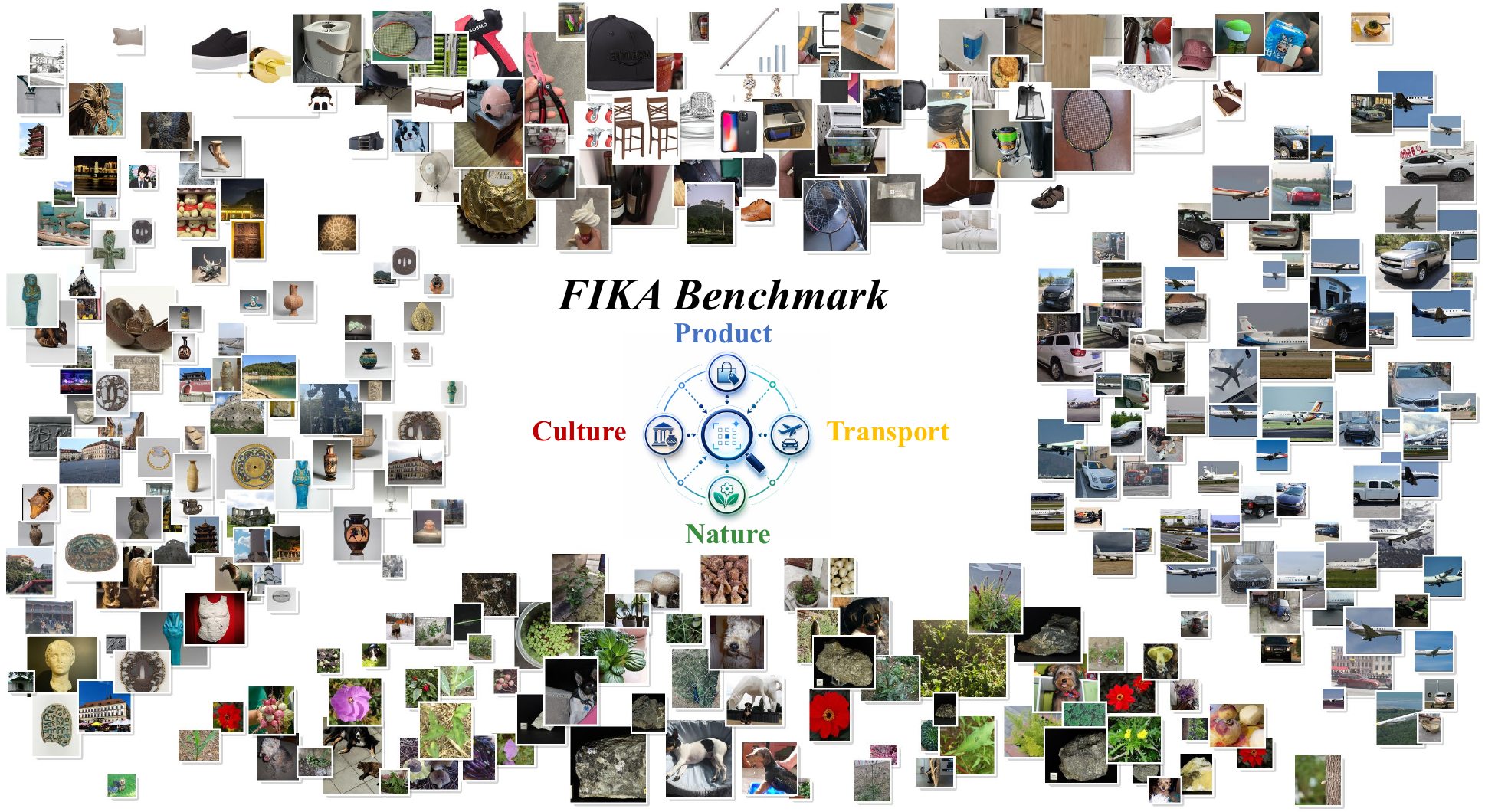}
        \caption{Overview of \benchmark{}: 311 evidence-grounded fine-grained instances from public-source and real-life images across Product, Nature, Transport, and Culture, evaluating whether models can acquire external fine-grained knowledge to recognize unseen categories.}
        \label{fig:dataset-instances}
    \end{minipage}
\end{figure*}

\begin{abstract}
    Fine-grained recognition in everyday life is often not a closed-book classification problem: when encountering unfamiliar objects, humans actively search, compare visual details, and verify evidence before deciding. Existing benchmarks primarily evaluate visually recognition, leaving this active external knowledge acquisition ability underexplored. We study \emph{fine-grained knowledge acquisition}, where a system must seek, verify, and use external evidence to answer open-ended fine-grained recognition questions. We introduce \benchmark{}, a leakage-aware and evidence-grounded collection of 311 public-source and real-life instances. To ensure high quality, every example is filtered against frontier closed-book models to remove memorized cases and audited to eliminate image-answer leakage, retaining only samples supported by verified evidence. Our evaluation of latest Large Multimodal Models (LMMs) and agents reveals that the task remains a formidable challenge: the best system reaches only 25.1\% accuracy, with no model exceeding 30\%. Crucially, we find that merely equipping models with tools is insufficient to bridge this gap; agent failures are predominantly driven by wrong entity retrieval and poor visual judgement. These results show that reliable knowledge acquisition needs better agent designs that focus on fine-grained recognition.
\end{abstract}
\section{Introduction}

Fine-grained recognition (FGR)~\citep{wah2011cub,krause2013cars,maji2013aircraft,horn2018inat} refers to visually identifying objects' specific categories instead of general categories, such as ``Snow Goose'' for a bird or ``BMW 530Li'' for a car. Fine-grained recognition is a fundamental capability used by humans every day to perceive and understand the world.
However, people do not know all fine-grained categories knowledge when they need to recognize them. When facing an unfamiliar vehicle, plant, product, or artifact, a person may inspect the logo, read a tail marking, take a photo, search the web, compare candidates against reference images, and use the collected evidence to justify a final decision.
In cognitive science and psychology, this capability is commonly referred to as \emph{epistemic action}~\citep{kirsh1994epistemic} and \emph{information foraging}~\citep{pirolli1999informationforaging}, which describe external actions taken to reveal hidden information or reduce uncertainty and how humans adapt strategies to obtain extra information from the environment.
We use \emph{Fine-grained knowledge acquisition} to refer to this ability in fine-grained recognition: a system should be able to seek and verify external evidence when the answer is not available with internal knowledge.

Early FGR benchmarks established the closed-set formulation for visually similar subordinate categories, such as bird species in CUB-200-2011 and NABirds, vehicle variants in Stanford Cars and FGVC-Aircraft, and dog breeds in Stanford Dogs \citep{wah2011cub,vanhorn2015nabirds,krause2013cars,maji2013aircraft,khosla2011dogs}.
This closed candidate-set assumption is useful for controlled evaluation, but it often fails in real-world recognition, where the relevant fine-grained label may be unknown.
Later datasets expanded fine-grained recognition to broader and more domain-specific taxonomies, including Food-101, VegFru, iNaturalist, Products-10K, and Google Landmarks v2 \citep{bossard2014food,hou2017vegfru,horn2018inat,bai2020products10k,weyand2020gldv2}. More recent LMM-era evaluations revisit fine-grained recognition through open-ended multimodal models and natural-language interfaces \citep{kim2024finer,yu2025fgbmk,pang2025frow,zhou2026worldvqa}. This progression made fine-grained recognition increasingly measurable, scalable, and multimodal, but the standard evaluation target remains largely passive: given an image and a label space or prompt, the model predicts an answer from its visual input and parametric knowledge.

This leaves four limitations for evaluating multimodal systems on fine-grained knowledge acquisition.

\begin{list}{$\bullet$}{
\setlength{\leftmargin}{1.2em}
\setlength{\labelwidth}{0.7em}
\setlength{\labelsep}{0.5em}
\setlength{\itemsep}{2pt}
\setlength{\topsep}{2pt}
\setlength{\parsep}{0pt}}
\item \textbf{Parametric saturation.} Many public fine-grained categories are no longer sufficiently difficult for frontier LMMs. If a model already internalizes the category, success measures recall rather than active recognition.
\item \textbf{Label granularity mismatch.} Fine-grained labels inherited from closed-set datasets can be coarse, ambiguous, or wrong for open-ended recognition. For example, visually similar aircraft variants such as Boeing 737-900 and 737-900ER are considered same in~\cite{maji2013aircraft}.
\item \textbf{Source-dataset leakage.} Public images may already be indexed online together with dataset pages, filenames, class names, or metadata. If a system can recover the answer by tracing the original dataset record, the benchmark measures leakage exploitation rather than transferable knowledge acquisition.
\item \textbf{Missing evidence support.} Existing datasets usually provide labels but not external evidence showing that the answer is verifiable and reachable through knowledge acquisition. Without such evidence, it is unclear whether a correct answer reflects supported recognition, unsupported guessing, or an unverifiable annotation.
\end{list}

To address these limitations, we present \benchmark{} (\emph{Fine-grained Knowledge Acquisition Benchmark}), a collection of 311 meticulously curated examples spanning 4 top-level and 17 mid-level categories. The construction of \benchmark{} follows a rigorous pipeline: \textbf{public-source screening} integrates diverse fine-grained recognition scenarios, \textbf{model-hard selection} filters out cases already solved by frontier LMMs such as Gemini-3 and GPT-5.4, \textbf{leakage inspection} removes direct image-answer shortcuts from source datasets, \textbf{label audit} corrects inherited annotations, and \textbf{evidence-grounded filtering} ensures every retained instance is supported by verifiable external evidence. This rigorous filtering leads to a high rejection rate. For instance, in the traditional FGVC collection, starting from 9,110 images across 911 classes, only 97 examples (roughly 1\%) survived our filters. This results in a high-quality benchmark that forces models to move beyond parametric memorization and actively acquire reliable knowledge.

Our evaluation of closed-book models and tool-enabled agents reveals that \benchmark{} remains a formidable challenge. The best-performing system achieves only 25.1\% overall accuracy, with no model exceeding the 30\% threshold. Interestingly, we find that real-life images are not inherently more difficult than curated public-source samples, as agent systems can often leverage contextual cues and searchable visual details in real-world scenes. However, our analysis underscores that existing agent frameworks, such as OpenClaw and OpenCode, remain insufficient; agent failures are dominated by incorrect entity retrieval and visual grounding errors. These findings indicate that while tools provide a path to external information, true fine-grained knowledge acquisition requires significant advancements in the architectural design and functional capabilities of agents tailored for fine-grained recognition.

\section{Related Work}

\subsection{Large Multimodal Models}

Large multimodal models (LMMs) connect visual perception with language modeling, enabling open-ended image understanding, dialogue, OCR, grounding, and visual question answering. Early generalist models such as Flamingo and BLIP-2~\citep{alayrac2022flamingo,li2023blip2} showed that large language models can be adapted to visual inputs through cross-modal interfaces and frozen unimodal components. Instruction-tuned LMMs such as LLaVA and Qwen-VL further improved interactive visual reasoning, localization, and text-reading abilities \citep{liu2023llava,bai2023qwenvl}.

Despite this progress, recent studies show that LMMs remain unreliable when recognition requires subordinate-level distinctions or external world knowledge. Finer reports substantial degradation on fine-grained visual categorization and attributes failures to a modality gap between visual inputs and text-side concept knowledge \citep{kim2024finer}. WorldVQA isolates atomic visual world knowledge and shows that recognizing visual entities should be separated from complex downstream reasoning \citep{zhou2026worldvqa}. FG-BMK and FROW broaden this observation by evaluating LMMs across multiple fine-grained datasets and open-world recognition settings \citep{yu2025fgbmk,pang2025frow}.
But these benchmarks mostly evaluate what the model already knows from its parameters and visual input. \benchmark{} instead asks whether a system can actively acquire missing category knowledge after examples solvable by strong closed-book LMMs have been filtered out.

\begin{figure}[!t]
  \centering
  \includegraphics[width=\linewidth]{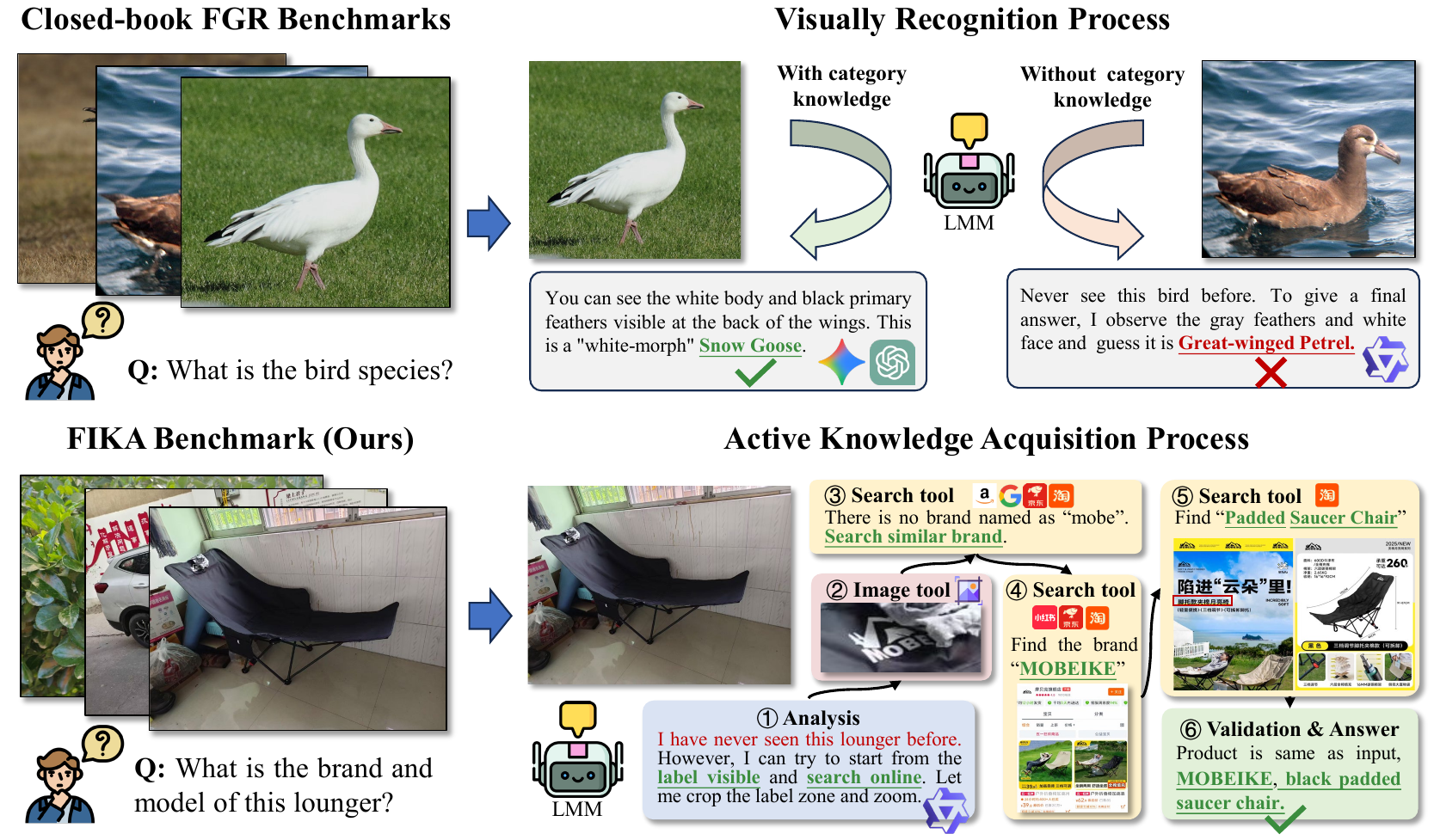}
  \caption{Comparison between \benchmark{} and existing fine-grained recognition benchmarks. \benchmark{} evaluates evidence-grounded fine-grained knowledge acquisition rather than only closed-set or open-ended recognition from the image and model parameters.}
  \label{fig:benchmark-comparison}
\end{figure}

\begin{table}[t]
\centering
\caption{\benchmark{} statistics. \textbf{Left:} source, language, and evidence composition. \textbf{Right:} semantic taxonomy, fine-grained answer coverage, and evidence-source diversity.}
\label{tab:dataset-statistics}
\small
\setlength{\tabcolsep}{8pt}
\scalebox{0.95}{
\begin{tabular}{lr|lr}
\toprule
\multicolumn{2}{c|}{\textbf{Dataset Composition}} & \multicolumn{2}{c}{\textbf{Semantic Taxonomy}} \\
\cmidrule(r){1-2} \cmidrule(l){3-4}
\textbf{Metric} & \textbf{Value} & \textbf{Statistic} & \textbf{Value} \\
\midrule
\textbf{Total Samples} & \textbf{311} & \multicolumn{2}{l}{\textbf{Top-level Category}} \\
 & & Culture (4) & 29.90\% \\
\multicolumn{2}{l|}{\textbf{Source Split}} & Transport (4) & 25.72\% \\
- Public datasets & 195 / 62.70\% & Nature (3) & 23.15\% \\
- Real-life collection & 116 / 37.30\% & Product (6) & 21.22\% \\
\multicolumn{2}{l|}{\textbf{Language}} & \multicolumn{2}{l}{\textbf{Fine-grained Answers}} \\
- English & 182 / 58.52\% & Unique answers & 228 \\
- Chinese & 129 / 41.48\% & Avg. samples / answer & 1.36 \\
\multicolumn{2}{l|}{\textbf{Evidence}} & \multicolumn{2}{l}{\textbf{Evidence Sources}} \\
- With evidence URL & 311 / 100.00\% & Evidence URLs & 319 \\
- Avg. URLs / sample & 1.03 & Unique domains & 120 \\
\bottomrule
\end{tabular}
}
\end{table}

\subsection{Multimodal Large-Model Agents}

Multimodal agents extend LMMs from passive perception to active problem solving. They can plan actions, browse webpages, crop or zoom images, perform OCR, retrieve external evidence, and revise intermediate hypotheses~\citep{he2024webvoyager,koh2024visualwebarena,zhang2024mmina}.

More recent systems emphasize search and deep-research behavior. MMSearch introduces an LMM-based multimodal search pipeline with query reformulation, reranking, summarization, and end-to-end search \citep{jiang2024mmsearch}. WebWatcher and BrowseComp-VL move toward vision-language deep research, where agents must combine visual anchors with complex web retrieval \citep{geng2025webwatcher}. These systems establish the feasibility and difficulty of multimodal tool use.

However, agent capability is usually measured by broad task success: completing a website workflow, finding an obscure event, or synthesizing a multimodal answer. \benchmark{} narrows this setting to a controlled scientific question: when a fine-grained visual category is unknown to the base model, can an agent obtain reliable external evidence and use it to make the correct recognition decision?

\subsection{Fine-Grained Classification Benchmarks}

Fine-grained visual classification (FGVC) benchmarks have long studied recognition among visually similar subordinate categories. Canonical datasets include CUB-200-2011 and NABirds for birds, Stanford Cars and FGVC-Aircraft for vehicles, and iNaturalist for large-scale species recognition \citep{wah2011cub,vanhorn2015nabirds,krause2013cars,maji2013aircraft,horn2018inat}. Domain-specific datasets such as Food-101, VegFru, and Products-10K extend fine-grained recognition to food, vegetables/fruits, and SKU-level products \citep{bossard2014food,hou2017vegfru,bai2020products10k}. These datasets made fine-grained recognition measurable and encouraged methods that exploit parts, attributes, hierarchies, and long-tail taxonomies.

Nevertheless, most FGVC benchmarks remain closed-set and public-data based. They typically provide a label but not evidence for why the label is correct; they also do not control whether a web-enabled agent can directly retrieve the original image and answer. \benchmark{} is designed to complement them by combining decontaminated public samples with real-life images, filtering closed-book successes, and attaching human-verified evidence to each fine-grained label.

\section{\benchmark}

\subsection{Problem Definition}

\benchmark{} tests whether a multimodal system can acquire external fine-grained knowledge rather than only recall categories knowledge already known. Formally, the benchmark is a set
\[
\mathcal{D}=\{(x_i,q_i,y_i,\mathcal{E}_i)\}_{i=1}^{N},
\]
where $x_i$ is the query image, $q_i$ is an open-ended fine-grained question, $y_i$ is the verified target answer, $\mathcal{E}_i$ is a set of human-collected evidence sources supporting the answer. Given $(x_i,q_i)$ and an optional tool set $\mathcal{T}$, a system $\pi$ outputs an answer $\hat{y}_i=\pi(x_i,q_i;\mathcal{T})$ and may optionally return supporting evidence $\hat{\mathcal{E}}_i$. Closed-book models are evaluated with $\mathcal{T}=\emptyset$, while multimodal agents may use search, browsing, or visual comparison tools.
The primary metric is strict answer accuracy:
\[
\mathrm{Acc}(\pi)=\frac{1}{N}\sum_{i=1}^{N}\mathbb{I}\left[\mathrm{match}(\hat{y}_i,y_i)=1\right],
\]
where $\mathrm{match}(\cdot)$ accepts aliases only when they preserve the required fine-grained granularity. Answers that name only a parent category, a visually similar sibling, or an unsupported candidate are counted as incorrect.

\subsection{Design Principles and Criteria}

\benchmark{} is guided by three strictly discard-oriented criteria.

\textbf{Active-knowledge necessity.} A candidate should not remain if strong LMMs can answer it reliably. Otherwise, knowledge acquisition capability would be obscured.

\textbf{Leakage resistance.} Direct online image-answer leakage is rigorously removed to prevent the inclusion of samples that a system could exploit to obtain ground-truth labels by tracing original datasets, thereby leading to metric hacking.

\textbf{Evidence-grounded fidelity.} Each answer must be supported by external evidence or verifiable visual cues, rather than only by legacy dataset labels. Otherwise, it would be impossible to rigorously validate the accessibility and utility of external fine-grained knowledge.

These principles distinguish \benchmark{} from conventional fine-grained classification benchmarks. Existing datasets often provide fixed category labels, but they do not ask whether a model can acquire external knowledge when the label is unknown, nor do they consistently expose whether a label is supported by reliable evidence. \benchmark{} therefore treats evidence and leakage status as part of the benchmark construction, not as post-hoc annotation metadata.

\subsection{Data Sources and Taxonomy}

\noindent\textbf{Public-source partition.}
The public partition contains 195 examples selected from existing fine-grained and open visual datasets, including FGVC-Aircraft, Stanford Cars, Stanford Dogs, Oxford Flowers-102, Food-101, VegFru, Google Landmarks v2, mineral images, The Met Open Access records, and Amazon Berkeley Objects \citep{maji2013aircraft,krause2013cars,khosla2011dogs,nilsback2008flowers,bossard2014food,hou2017vegfru,weyand2020gldv2,mineralimage5k98,metmuseumopenaccess,collins2022abo}. We do not directly reuse these datasets as-is. Instead, public-source candidates are screened for model hardness, direct image-answer leakage, label correctness, and evidence availability before entering \benchmark{}.

\noindent\textbf{Real-life partition.}
The real-life partition contains 116 images collected from volunteers' daily scenarios, such as products, vehicles, plants, landmarks, venues, and cultural objects encountered outside curated benchmark settings. These examples complement public datasets by reflecting the open-world situations in which a user may need to inspect visual details, search external sources, and verify a fine-grained answer from evidence.

\noindent
Both partitions are mapped to the same two-level taxonomy: Product, Nature, Transport, and Culture as top-level categories, with 17 mid-level categories shared across the full benchmark. This design lets us compare controlled public-source recognition with more realistic in-the-wild knowledge acquisition under a unified semantic taxonomy. Figure~\ref{fig:taxonomy-radar} visualizes the resulting two-level distribution.

\begin{table*}[t]
\centering
\small
\caption{Comparison of different systems on \benchmark. We report fine-grained classification accuracy across public-dataset and real-life splits using the same category taxonomy: Prod., Nat., Trans., and Cult. denote Product, Nature, Transport, and Culture. \textbf{Bold} and \underline{underline} indicate the best and second-best results, respectively.}
\label{tab:main-results}
\resizebox{\textwidth}{!}{
\begin{tabular}{lccccccccccc}
    \toprule

    \multirow{2}{*}{\textbf{System}}
    & \multicolumn{5}{c}{\textbf{Public}}
    & \multicolumn{5}{c}{\textbf{Real-Life}}
    & \multirow{2}{*}{\textbf{Overall}} \\
    \cmidrule(lr){2-6} \cmidrule(lr){7-11}
    & \textbf{Prod.} & \textbf{Nat.} & \textbf{Trans.} & \textbf{Cult.} & \textbf{Avg.}
    & \textbf{Prod.} & \textbf{Nat.} & \textbf{Trans.} & \textbf{Cult.} & \textbf{Avg.}
    & \\
    \midrule
    \multicolumn{12}{l}{\textit{\textbf{Closed-Source Models}}} \\
    GPT-5-mini~\citep{openai2025gpt5mini}
        & 3.7 & 12.0 & \underline{41.1} & 0.0 & 15.4
        & 15.4 & \textbf{45.5} & 25.0 & 6.5 & 20.7 & 17.4 \\
    Gemini-3.1-Flash-Lite~\citep{deepmind2026gemini31flashlite}
        & 3.7 & \underline{20.0} & 37.5 & 1.6 & \underline{16.9}
        & \textbf{25.6} & 36.4 & \underline{29.2} & 19.4 & 26.7 & \underline{20.6} \\
    GLM-5V-Turbo~\citep{glmvteam2026glm5vturbo}
        & 3.7 & 16.0 & 23.2 & \underline{4.8} & 12.8
        & 10.3 & \underline{40.9} & \textbf{33.3} & 9.7 & 20.7 & 15.8 \\
    \midrule
    \multicolumn{12}{l}{\textit{\textbf{Open-Source and Fine-Grained Models}}} \\
    Kimi-K2.6~\citep{moonshot2026kimik26}
        & 0.0 & \textbf{26.0} & \textbf{44.6} & \textbf{6.5} & \textbf{21.5}
        & \underline{23.1} & \underline{40.9} & \textbf{33.3} & \textbf{32.3} & \textbf{31.0} & \textbf{25.1} \\
    Qwen3-VL-235B-A22B~\citep{bai2025qwen3vltechnicalreport}
        & \textbf{11.1} & 10.0 & 25.0 & 0.0 & 11.3
        & 12.8 & 31.8 & 20.8 & \underline{22.6} & 20.7 & 14.8 \\
    Qwen3.5-397B-A17B~\citep{qwen2026qwen35}
        & \textbf{11.1} & 12.0 & 28.6 & 3.2 & 13.8
        & 20.5 & 31.8 & \textbf{33.3} & \underline{22.6} & 25.9 & 18.3 \\
    Qwen3-VL-8B~\citep{bai2025qwen3vltechnicalreport}
        & 0.0 & 6.0 & 17.9 & 0.0 & 6.7
        & 7.7 & 27.3 & 12.5 & 9.7 & 12.9 & 9.0 \\
    Qwen3.5-9B~\citep{qwen2026qwen35}
        & 0.0 & 2.0 & 17.9 & 0.0 & 5.6
        & 7.7 & \textbf{45.5} & 12.5 & 6.5 & 15.5 & 9.3 \\
    Fine-R1-7B~\citep{he2026finer1}
        & 3.7 & 14.0 & 21.4 & 3.2 & 11.3
        & 7.7 & 18.2 & 16.7 & 9.7 & 12.1 & 11.6 \\
    VisualRFT-7B~\citep{liu2025visualrft}
        & 3.7 & 8.0 & 17.9 & 0.0 & 7.7
        & 5.1 & 4.5 & 4.2 & 6.5 & 5.2 & 6.8 \\
    \midrule
    \multicolumn{12}{l}{\textit{\textbf{Multimodal Agents}}} \\
    OpenClaw + Qwen3-VL-8B~\citep{openclaw2026openclaw,bai2025qwen3vltechnicalreport}
        & 0.0 & 4.0 & 25.0 & 0.0 & 8.2
        & 20.5 & \underline{40.9} & 16.7 & 19.4 & 23.3 & 13.8 \\
    OpenClaw + Qwen3.5-397B-A17B~\citep{openclaw2026openclaw,qwen2026qwen35}
        & \underline{7.4} & 16.0 & 25.0 & 3.2 & 13.3
        & \underline{23.1} & \underline{40.9} & \textbf{33.3} & \underline{22.6} & \underline{28.4} & 19.0 \\
    OpenClaw + MiniMax-M2.7/Qwen3-VL-8B~\citep{openclaw2026openclaw,minimax2026m27,bai2025qwen3vltechnicalreport}
        & 0.0 & 6.0 & 21.4 & 0.0 & 7.7
        & 17.9 & 27.3 & 25.0 & 16.1 & 20.7 & 12.5 \\
    OpenCode + Qwen3-VL-8B~\citep{opencode2026opencode,bai2025qwen3vltechnicalreport}
        & 3.7 & 4.0 & 19.6 & 1.6 & 7.7
        & 5.1 & 22.7 & 16.7 & 0.0 & 9.5 & 8.4 \\
    OpenCode + MiniMax-M2.7/Qwen3-VL-8B~\citep{opencode2026opencode,minimax2026m27,bai2025qwen3vltechnicalreport}
        & 0.0 & 12.0 & 16.1 & 0.0 & 7.7
        & 7.7 & 27.3 & 20.8 & 9.7 & 14.7 & 10.3 \\
    \bottomrule
\end{tabular}
}
\end{table*}

\subsection{Construction Pipeline}

\textbf{Step 1: Candidate pool construction.} We first built broad candidate pools rather than directly sampling final evaluation items. For traditional FGVC sources, we evaluated six datasets with usable image-level fine-grained labels: FGVC-Aircraft, Stanford Cars, Stanford Dogs, Oxford Flowers-102, Food-101, and VegFru. Using Qwen3-VL-8B-Instruct-FP8 in a no-candidate closed-book protocol, we screened 911 classes and 9,110 images. This produced 282 low-accuracy classes and 2,820 candidate images for the next stage. In parallel, we explored additional public sources for general domain coverage, including minerals, landmarks, museum artifacts, daily objects, fashion/accessories, product packaging, and difficult food. The full source list and its use in the final benchmark are summarized in Appendix Table~\ref{tab:source-usage}.

\textbf{Step 2: Model-hard candidate selection.} To reduce parametric-knowledge shortcuts, we applied model-based filtering before human audit. GPT-5.4-mini and Gemini-3-Flash were evaluated on samples from the previous stage. We retained classes whose closed-book accuracy was below 40\% and only kept the wrong samples, yielding 435 hard candidates from 52 classes. This class-level rule avoids retaining isolated failures while keeping categories where strong models fail consistently.

\textbf{Step 3: Leakage inspection.} Candidates are filtered for direct image-answer leakage before label review. We use an image-copy detection pipeline inspired by ISC-style copy detection \citep{douze2021isc}: candidate duplicates are retrieved from web-scale image search results and then verified by near-duplicate similarity. Samples are removed when the image is leaked by the source dataset in a way that makes the answer directly recoverable, or when leakage status cannot be determined.

\textbf{Step 4: Label audit.} After leakage inspection, human annotators verify whether the inherited dataset label is specific enough for an open-ended fine-grained answer. Crucially, we found this step to be indispensable: in the traditional FGVC branch, 257 labels were corrected out of 353 cleaned samples. This is largely because public FGR benchmarks are often tailored for closed-set settings, resulting in labels that are either ambiguous or insufficiently granular. For instance, variants like the Boeing 737-900 and 737-900ER are often treated as a single class in standard aircraft datasets for simplicity, despite being distinct models. Samples with uncertain labels are dropped rather than retained with ambiguous ground truth.

\textbf{Step 5: Evidence review and filtering.} Annotators then collect evidence for the corrected answer. We do not restrict which search engine, visual search tool, marketplace, encyclopedia, museum page, or social platform may be used during annotation, but every retained evidence item must be stored as an accessible URL in the released record. Samples without usable evidence are filtered out. The final benchmark contains 319 evidence URLs from 120 unique domains. These include official sources such as \texttt{metmuseum.org}, product and marketplace pages such as \texttt{item.jd.com}, \texttt{detail.tmall.com}, and Amazon-region pages, taxonomic or encyclopedic sources such as \texttt{baike.baidu.com} and Wikipedia, aviation and vehicle databases such as \texttt{jetphotos.com} and \texttt{planespotters.net}, and tool or social sharing links such as \texttt{xhslink.com}.

\subsection{Evaluation Protocol}

We report answer accuracy as the primary metric. A prediction is correct only if it matches the required fine-grained answer or an accepted alias at the target granularity. Overly broad answers, such as naming only the parent class or silbing class are counted as incorrect.
We implement this rule with a strict LLM-as-judge protocol. The judge receives the question, the verified answer, and the model prediction, and returns a verdict from \texttt{correct}, \texttt{partially\_correct}, \texttt{incorrect}, or \texttt{uncertain}. Only \texttt{correct} is counted in the main accuracy. Appendix~\ref{app:judge-prompt} reports the exact judge prompt.

\section{Experiments}

\subsection{Settings}

We mainly evaluate two paradigms: closed-book models, which lack external information access during inference, and multimodal agents capable of invoking external tools. All closed-book models, including fine-grained-specialized models, receive identical image-question prompts without candidate labels. Agents receive the same prompts, supplemented by temporary image paths for tool-based local file access. Detailed prompt templates and agent toolsets are provided in Appendices~\ref{app:closed-book-prompt} and~\ref{app:agent-task-prompt}. To ensure efficiency and result independence, we impose a 240-second timeout per example and disable cross-sample memory inheritance.

\subsection{Main Results}

\textbf{Current systems remain far from reliable fine-grained knowledge acquisition.} As shown in Table~\ref{tab:main-results}, the strongest candidate is the open-source Kimi-K2.6, with an overall accuracy of 25.1\%, followed by Gemini-3.1-Flash-Lite at 20.6\% and OpenClaw + Qwen3.5-397B-A17B at 19.0\%. No system reaches 30\% overall accuracy, even though the benchmark contains human-verified answers with accessible evidence. This indicates that \benchmark{} is substantially harder than conventional FGVR evaluations testing closed-set visual discrimination, because the model must recover open-ended fine-grained knowledge and match the required answer specificity.

\textbf{The public-source partition remains difficult despite originating from curated datasets.} The overall best-performing model, Kimi-K2.6, achieves only an average score of 21.5\%, while Qwen3.5-397B-A17B reaches 13.8\%. The Public Product and Culture categories are particularly challenging: most systems score near zero on Product, and no evaluated system exceeds 6.5\% on Culture. Public Transport is comparatively easier, with Kimi-K2.6 reaching 44.6\%, suggesting that aircraft and vehicle cues are more recoverable from model priors or visually distinctive markings than product brands or cultural artifacts.

\textbf{The real-life partition shows a different error profile.} Several models perform better on real-life images than on public-source samples, with Kimi-K2.6 reaching 31.0\%, OpenClaw + Qwen3.5-397B-A17B reaching 28.4\%, and Gemini-3.1-Flash-Lite reaching 26.7\% on the real-life average. The shift is especially visible for agent systems: OpenClaw + Qwen3.5-397B-A17B improves from 13.3\% on the public average to 28.4\% on the real-life average, and OpenClaw + Qwen3-VL-8B improves from 8.2\% to 23.3\%. This indicates that the agent paradigm can be more adaptive for real-world fine-grained recognition, where contextual cues and searchable visual details are often available.

\textbf{Fine-grained-specialized models do not automatically transfer to this setting.} Fine-R1-7B reaches 11.6\% overall, while VisualRFT-7B reaches 6.8\%, both below several general-purpose models. This result is consistent with the design of \benchmark{}: success requires open-ended answer generation and evidence-aligned specificity, not only closed-set visual discrimination over a fixed label space.

\section{Analysis}

\begin{table*}[t]
\centering
\small
\caption{Agent tool-use statistics over 311 examples. Usage (\%) denotes the ratio of episodes with at least one tool call; Avg. Acts is the average number of actions per example.}
\label{tab:agent-tool-use}
\resizebox{\textwidth}{!}{
\begin{tabular}{lcccc}
\toprule
\textbf{System} & \textbf{Usage (\%)} & \textbf{Avg. Acts} & \textbf{Tool Err (\%)} & \textbf{Tool Usage Breakdown (Count)} \\
\midrule
OpenClaw + Qwen3-VL-8B               & 97.7 & 1.94 & 90.1 & image inspection 272; reverse image search 242 \\
OpenClaw + MiniMax-M2.7/Qwen3-VL-8B  & 99.7 & 1.66 & 6.0  & image inspection 322; shell/exec 127 \\
OpenCode + Qwen3-VL-8B               & 47.9 & 0.51 & 1.2  & reverse image search 152 \\
OpenCode + MiniMax-M2.7/Qwen3-VL-8B  & 92.0 & 0.92 & 0.0  & image inspection 287 \\
\bottomrule
\end{tabular}}
\end{table*}

\paragraph{Agent tool-use suitability.}
We further analyze how current agents behave and use tools in fine-grained recognition. The analysis in Table~\ref{tab:agent-tool-use} covers the four Qwen3-VL-8B-based agent settings, with tool and skill details reported in Appendix~\ref{app:agent-tools}.
Table~\ref{tab:agent-tool-use} shows that all four systems rely primarily on visual evidence acquisition, but they expose very different tool-use patterns. The two OpenClaw settings invoke actions on almost every example, with 97.7\%--99.7\% action-use rates and 1.66--1.94 action calls per example. OpenCode+MiniMax-M2.7/Qwen3-VL-8B also uses tools frequently, while OpenCode+Qwen3-VL-8B is much more conservative, using actions on 47.9\% of examples with only 0.51 calls per example. Across systems, the dominant calls are image inspection or reverse image search, indicating that agents mostly spend their external-action budget on extracting or matching visual evidence rather than on broad web browsing.

The table distinguishes between tool invocation frequency and successful execution. While OpenClaw+Qwen3-VL-8B exhibits the highest action intensity, it suffers from a 90.1\% tool-error rate. Trace analysis reveals that most failures (519 of 544) are ``tool not found'' errors rather than retrieval failures, primarily due to attempts to call unavailable tools (e.g., \texttt{image}) or treat skills like \texttt{reverse-image-search} as direct tool calls. Conversely, MiniMax-M2.7/Qwen3-VL-8B variants utilize explicit image-inspection command, and OpenCode routes skills via shell execution. These results suggest that agent performance depends not only on tool availability but also on the model's ability to align its behavior with the framework.

\begin{figure}[t]
  \centering
  \includegraphics[width=0.882\linewidth]{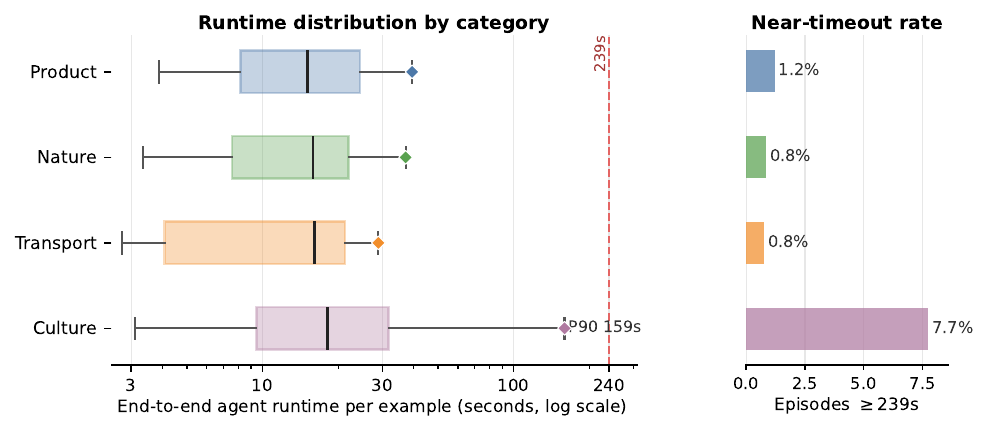}
  \caption{End-to-end agent runtime distribution by top-level category. Boxes show the interquartile range, whiskers show P10--P90, diamonds mark P90, and the dashed line marks the 239s near-timeout threshold.}
  \label{fig:agent-runtime-distribution}
\end{figure}

\paragraph{Runtime distribution.}
We also examine end-to-end agent runtime distribution. Figure~\ref{fig:agent-runtime-distribution} shows that Product, Nature, and Transport have similar median runtimes of 15.1s, 15.9s, and 16.1s, with 90th-percentile runtimes of 39.4s, 37.1s, and 28.9s. Culture has a comparable median of 18.1s but a much heavier tail: its 90th percentile reaches 159.4s, and 7.7\% of Culture episodes run into the near-timeout region of \(\geq 239\)s, compared with at most 1.2\% for any other top-level category. The runtime profile therefore suggests category-dependent operating modes: Product, Nature, and Transport are often resolved through relatively short visual matching or targeted lookup, whereas Culture more often triggers longer evidence search and disambiguation.

\paragraph{Memory and targeted skill evolution.}
We next investigate two potential pathways for enhancing agent performance: maintaining memory across examples and developing task-specific skills. We study this on the Aircraft subset, where many airplane images contain a registration mark available in public aviation databases, but is inaccessible to closed-book models at inference time. Memory inheritance improves strict accuracy from 20.9\% to 30.2\% by reusing verified operators, aircraft families, and visual cues across examples.

\begin{table}[t]
\centering
\small
\caption{Aircraft subset comparison for memory inheritance and targeted skill evolution. All rows use OpenClaw+Qwen3.5-397B-A17B on the same 43 Aircraft examples. Strict Acc. counts only correct responses; Soft Acc. incorporates both correct and partially correct results.}
\label{tab:aircraft-memory-skill}
\resizebox{0.75\linewidth}{!}{
\begin{tabular}{lrrrrr}
\toprule
\textbf{Mode} & \textbf{Correct} & \textbf{Partial} & \textbf{Incorrect} &  \textbf{Strict Acc.} & \textbf{Soft Acc.} \\
\midrule
Baseline & 9 & 5 & 28 &  20.9 & 32.6 \\
Memory inheritance & 13 & 3 & 26 &  30.2 & 37.2 \\
Registration skill & 15 & 3 & 18 & 34.9 & 41.9 \\
\bottomrule
\end{tabular}}
\end{table}

Moreover, a targeted skill provides a larger gain than memory alone. As summarized in Table~\ref{tab:aircraft-memory-skill}, adding an aircraft-registration lookup skill raises the same no-memory setting to 15 correct and 3 partially correct answers over the 43 Aircraft examples. The correct rate increases from 20.9\% without memory to 30.2\% with memory, and to 34.9\% with the specialized skill. This suggests that current fine-grained agents still have substantial room to improve through memory, task-specific skills, and self-evolution, rather than relying only on stronger backbone models.

\begin{figure}[t]
  \centering
  \includegraphics[width=0.92\linewidth]{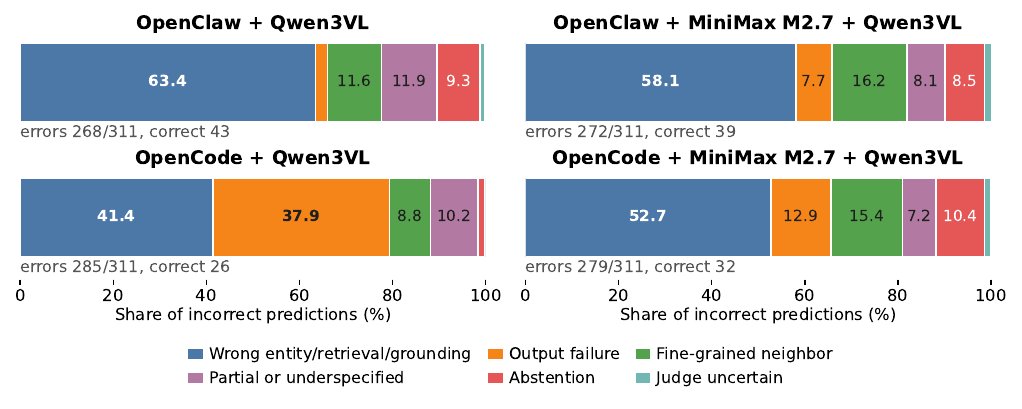}
  \caption{Error taxonomy for current agent methods. Each panel reports the distribution over non-strict-correct predictions for one agent configuration.}
  \label{fig:agent-error-taxonomy}
\end{figure}

\paragraph{Agent error taxonomy.}
\label{sec:agent-error-taxonomy}
To understand the primary limitations of existing agents in fine-grained recognition, we conduct an error taxonomy analysis of incorrect agent predictions. This analysis covers the four Qwen3-VL-8B-based agent settings reported in Table~\ref{tab:main-results}.
Figure~\ref{fig:agent-error-taxonomy} reveals two prominent patterns. First, errors involving incorrect entities, retrieval, or visual grounding dominate every setting, ranging from 41.4\% for OpenCode+Qwen3-VL-8B to 63.4\% for OpenClaw+Qwen3-VL-8B. Aggregated across systems, these categories account for 53.7\% of all agent failures. Within this error class, agents often fail to retrieve the correct evidence or struggle to distinguish valid evidence from similar distractors, ultimately leading to incorrect judgments. This underscores that there is still substantial room for improvement in agents specialized for fine-grained recognition.
Second, the reliability profile differs substantially by framework. OpenCode+Qwen3-VL-8B has a large generation or empty-output block, accounting for 37.9\% of its errors, whereas the same category is only 2.6\% for OpenClaw+Qwen3-VL-8B. Adding MiniMax-M2.7 as the controller reduces OpenCode output failures to 12.9\%, but shifts more remaining errors into semantic categories such as wrong-entity retrieval and fine-grained neighbor confusion.

\section{Conclusion}

We introduced \benchmark{}, a leakage-aware and evidence-grounded benchmark for fine-grained knowledge acquisition. Unlike conventional fine-grained recognition benchmarks that primarily evaluate visually recognition from an image or fixed prompt, \benchmark{} asks whether a system can seek, verify, and use external evidence when internal knowledge is insufficient.
Our evaluation shows that this capability remains far from solved. The best system reaches only 25.1\% overall accuracy, and no evaluated model or agent exceeds 30\%. Tool-enabled agents also do not automatically bridge the gap: their failures are dominated by incorrect entity retrieval and visual grounding errors. These results suggest that reliable fine-grained knowledge acquisition requires more than tool access; it requires stronger evidence verification, exact entity disambiguation, and agent mechanisms better tailored to fine-grained recognition.

\paragraph{Limitations and future work.}
\benchmark{} is designed as a high-precision, evidence-grounded benchmark, so the current release prioritizes annotation reliability and leakage control over exhaustive coverage of every fine-grained domain. As models and web content evolve, the dataset taxonomy, evidence links, and product or place categories should be periodically refreshed. Future versions can also broaden multilingual and regional coverage and include richer process-level metrics for agent behavior, while keeping the same human-audited protocol.

\bibliographystyle{plainnat}
\bibliography{references}

\clearpage
\appendix
\section{Additional Sample Visualizations}

\begin{figure}[!htbp]
  \centering
  \includegraphics[width=0.72\linewidth]{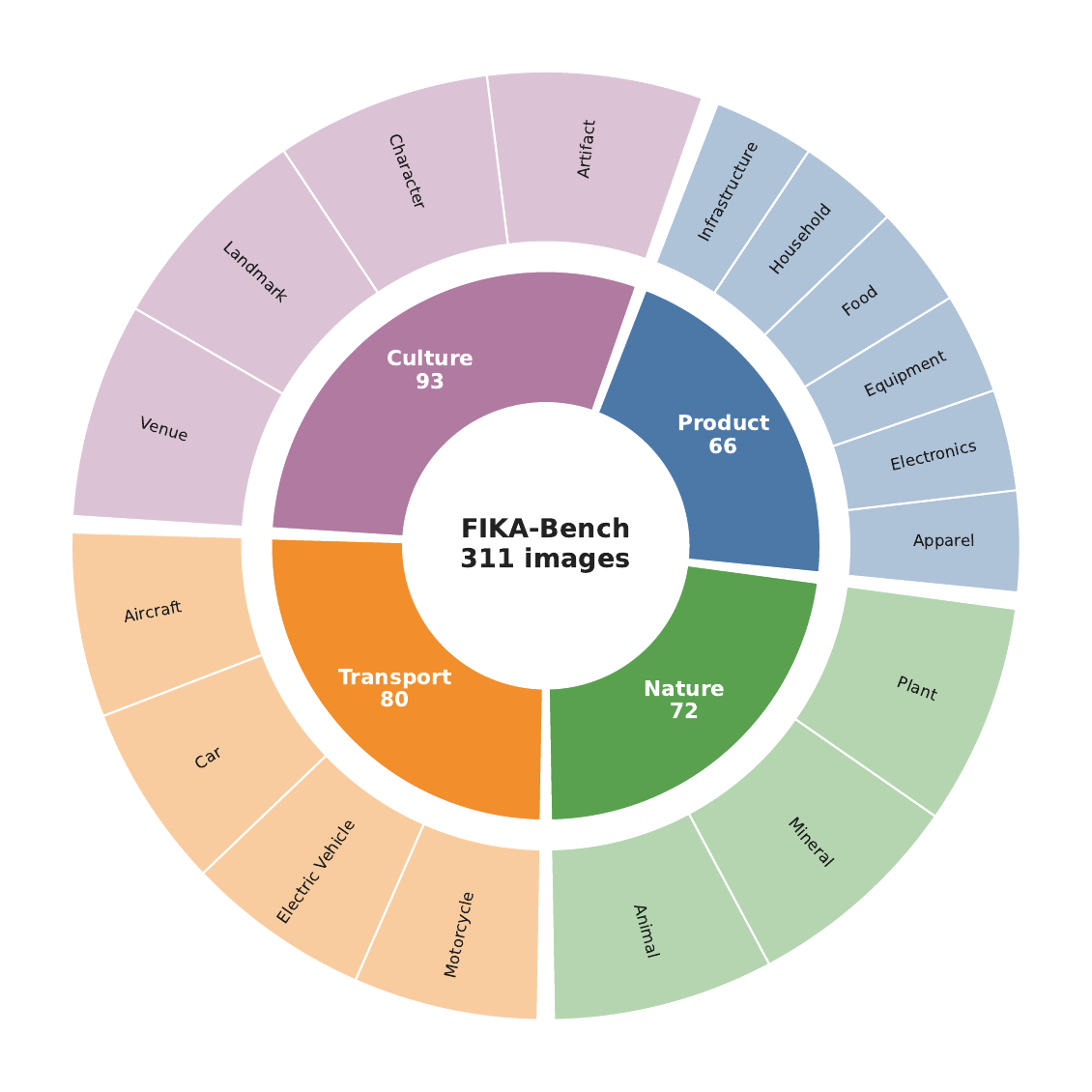}
  \caption{Two-level taxonomy distribution in \benchmark. Inner ring: top-level categories; outer ring: mid-level categories.}
  \label{fig:taxonomy-radar}
\end{figure}

\begin{figure}[!htbp]
  \centering
  \input{figures/appendix_category_samples}
  \caption{Representative \benchmark{} samples grouped by top-level category. Each row contains one public-data example and one real-life example from the same top-level category; each panel reports the top-level category, mid-level category, source partition, question, verified answer, and full evidence link.}
  \label{fig:appendix-category-samples}
\end{figure}

\section{Additional Benchmark Comparison}

\begin{table}[!htbp]
\centering
\caption{Comparison of \benchmark{} with representative benchmark families.}
\label{tab:benchmark-comparison}
\small
\setlength{\tabcolsep}{4pt}
\resizebox{\textwidth}{!}{
\begin{tabular}{lcccc>{\raggedright\arraybackslash}p{5.3cm}}
\toprule
\textbf{Benchmark family} & \textbf{Visual input} & \textbf{Agent tools} & \textbf{FG label} & \textbf{Leakage / closed-book filter} & \multicolumn{1}{c}{\textbf{Primary focus}} \\
\midrule
Classic FGVC \citep{wah2011cub,krause2013cars,maji2013aircraft} & Yes & No & Yes & No & Closed-set recognition over fixed fine-grained categories. \\
LMM FGVC evals \citep{kim2024finer,yu2025fgbmk,pang2025frow} & Yes & No & Yes & No & Fine-grained recognition and reasoning by LMMs. \\
Web/GUI agents \citep{he2024webvoyager,koh2024visualwebarena,xie2024osworld} & Yes & Yes & No & No & Website, browser, or computer task completion. \\
Multimodal browsing \citep{jiang2024mmsearch,tao2025mmsearchplus,wang2026merrin} & Yes & Yes & No & Partial & Retrieval and reasoning over multimodal web evidence. \\
\textbf{\benchmark{} (ours)} & Yes & Yes & Yes & Yes & Evidence-grounded active knowledge acquisition for fine-grained recognition. \\
\bottomrule
\end{tabular}
}
\end{table}

\begin{table*}[!htbp]
\centering
\small
\caption{Source datasets and their use in \benchmark{} construction. Original samples denote the source-specific pool that entered our scan or audit logs, not necessarily the full public dataset size. Final counts refer to retained samples in the final benchmark.}
\label{tab:source-usage}
\setlength{\tabcolsep}{4pt}
\begin{tabular}{@{}>{\raggedright\arraybackslash}p{0.26\linewidth}>{\raggedright\arraybackslash}p{0.17\linewidth}>{\raggedright\arraybackslash}p{0.10\linewidth}>{\raggedright\arraybackslash}p{0.40\linewidth}@{}}
\toprule
\textbf{Source} & \textbf{Original samples} & \textbf{Final kept} & \textbf{Use in construction} \\
\midrule
FGVC-Aircraft~\citep{maji2013aircraft} & 1,000 images / 100 classes & 42 & Vehicle/aircraft fine-grained examples after model-hard filtering, leakage inspection, label correction, and evidence review. \\
Stanford Cars~\citep{krause2013cars} & 1,960 images / 196 classes & 14 & Vehicle model examples; inherited labels are manually checked for open-ended answer granularity. \\
Stanford Dogs~\citep{khosla2011dogs} & 1,200 images / 120 classes & 13 & Animal examples retained only when strong closed-book models fail and external evidence supports the breed label. \\
Oxford Flowers-102~\citep{nilsback2008flowers} & 1,020 images / 102 classes & 6 & Plant examples retained after class-level hardness filtering and evidence review. \\
VegFru~\citep{hou2017vegfru} & 2,920 images / 292 classes & 22 & Plant and food examples retained from low-accuracy classes. \\
Food-101~\citep{bossard2014food} & 1,010 images / 101 classes & 0 & Screened for hard food examples but not retained in the final benchmark after filtering. \\
Google Landmarks v2~\citep{weyand2020gldv2} & 50 images / 10 classes & 19 & Landmark and venue examples after leakage and label-status review. \\
MineralImage5K98~\citep{mineralimage5k98} & 50 images / 10 classes & 9 & Mineral examples after leakage and label-status review. \\
The Met Open Access~\citep{metmuseumopenaccess} & 74 museum candidates & 43 & Culture/artifact examples backed by official Met records; model-correct samples are removed. \\
Amazon Berkeley Objects~\citep{collins2022abo} & 68 brand samples & 27 & Public Product supplement; samples are retained only when image leakage is not found, strong models fail, and brand evidence is verified. \\
Real-life collection & 166 annotated daily-life images & 116 & In-the-wild Product, Nature, Transport, and Culture examples after review, evidence collection, and privacy redaction. \\
THINGS~\citep{hebart2019things} & 1,854 concepts & 0 & Used to assess possible object-concept coverage; discarded because candidates were too easy for closed-book models. \\
Fashion Product Images~\citep{fashionproductimages} & 44,072 images & 0 & Used to assess fashion/product coverage; discarded because candidates were too easy, leaked, and unsuitable for evidence-grounded open-ended questions. \\
GroceryStoreDataset~\citep{klasson2019grocery} & 5,421 images & 0 & Used to assess grocery/product coverage; discarded because candidates were too easy for closed-book models. \\
UECFood-256~\citep{kawano2014uecfood256} & 31,395 images & 0 & Used to assess difficult food coverage; discarded because candidates were too easy for closed-book models. \\
\bottomrule
\end{tabular}
\end{table*}

\section{Agent Tool and Skill Configuration}
\label{app:agent-tools}

All agent systems receive the task template in Appendix~\ref{app:agent-task-prompt}. The benchmark question is kept fixed across agent settings; only the agent framework, controller model, visual access path, and available tools differ. Dataset labels, evidence URLs, source names, and category metadata are not provided to the agent.

\noindent\textbf{Agent configurations.}
\begin{list}{$\bullet$}{
\setlength{\leftmargin}{1.2em}
\setlength{\labelwidth}{0.7em}
\setlength{\labelsep}{0.5em}
\setlength{\itemsep}{2pt}
\setlength{\topsep}{2pt}
\setlength{\parsep}{0pt}}
    \item \textbf{OpenClaw + Qwen3-VL-8B.} The primary multimodal model is Qwen/Qwen3-VL-8B-Instruct-FP8. Enabled tools are \texttt{read}, \texttt{web\_fetch}, and \texttt{exec}. Enabled skills are \texttt{local-web-search} and \texttt{reverse-image-search}.
    \item \textbf{OpenClaw + MiniMax-M2.7/Qwen3-VL-8B.} The text controller is MiniMax-M2.7, with Qwen/Qwen3-VL-8B-Instruct-FP8 used as the image model. Enabled tools are \texttt{read}, \texttt{web\_fetch}, \texttt{image}, and \texttt{exec}. Enabled skills are \texttt{local-web-search} and \texttt{reverse-image-search}.
    \item \textbf{OpenCode + Qwen3-VL-8B.} We use opencode-ai 1.14.35 with Qwen/Qwen3-VL-8B-Instruct-FP8 through local vLLM, with image attachments enabled by model modality. Enabled tools are \texttt{read}, \texttt{webfetch}, \texttt{websearch}, \texttt{bash}, \texttt{edit}, \texttt{write}, \texttt{patch}, \texttt{multiedit}, \texttt{apply\_patch}, and \texttt{skill}. Enabled skills are \texttt{local-web-search} and \texttt{reverse-image-search}.
    \item \textbf{OpenCode + MiniMax-M2.7/Qwen3-VL-8B.} The text controller is MiniMax-M2.7, with Qwen/Qwen3-VL-8B-Instruct-FP8 exposed through a local \texttt{qwen\_image} inspection tool. Enabled tools are \texttt{read}, \texttt{webfetch}, \texttt{websearch}, \texttt{bash}, \texttt{edit}, \texttt{write}, \texttt{patch}, \texttt{multiedit}, \texttt{apply\_patch}, \texttt{skill}, and \texttt{qwen\_image}. Enabled skills are \texttt{local-web-search} and \texttt{reverse-image-search}.
\end{list}

\noindent\textbf{Installed skills.}
\begin{list}{$\bullet$}{
\setlength{\leftmargin}{1.2em}
\setlength{\labelwidth}{0.7em}
\setlength{\labelsep}{0.5em}
\setlength{\itemsep}{2pt}
\setlength{\topsep}{2pt}
\setlength{\parsep}{0pt}}
    \item \texttt{local-web-search}: a local DuckDuckGo HTML search wrapper that returns structured text-search results without a paid search API.
    \item \texttt{reverse-image-search}: a PicImageSearch-based reverse-image search skill using API-key-free engines, with Baidu searched first and TinEye second.
\end{list}

\section{Prompt Templates}

This appendix reports the prompt templates used for the main experiments and judging. Placeholders such as \texttt{<benchmark question>} and \texttt{<opaque local image path>} are replaced for each sample. Line wrapping is for typesetting only.

\subsection{Closed-Book VQA Prompt}
\label{app:closed-book-prompt}

Closed-book models and fine-grained-specialized models receive the image together with the following system and user messages. No candidate labels, evidence URLs, source dataset names, or ground-truth metadata are provided.

\begin{tcolorbox}[
title=Closed-Book VQA Prompt,
colback=white,
colframe=black!70,
arc=5pt,
fonttitle=\bfseries\ttfamily]

\textbf{System Message}

You are a visual question answering model for fine-grained recognition.
Answer the user's question using the image.
Return only a JSON object with these keys:
\begin{itemize}
    \item \texttt{answer}: the most specific answer supported by the image. Prefer the exact fine-grained category, entity, model, species, artifact title, venue, or geographic location when applicable; do not simplify a specific answer into a broader category.
    \item \texttt{confidence}: a number from 0 to 1
    \item \texttt{rationale}: one short sentence explaining the visual evidence
\end{itemize}

\vspace{4pt}
\textbf{User Message Template}

\begin{quote}
{\ttfamily\small
<image>\par
Question:\par
<benchmark question>\par
\par
Return only the JSON object. Do not include Markdown fences.\par
}
\end{quote}
\end{tcolorbox}

\subsection{Agent Task Prompt}
\label{app:agent-task-prompt}

Agent systems receive the same benchmark question. When a tool or skill requires a local image path, we provide an opaque temporary filename that contains no dataset or class information.

\begin{tcolorbox}[
title=Agent Task Prompt,
colback=white,
colframe=black!70,
arc=5pt,
fonttitle=\bfseries\ttfamily]

\begin{quote}
{\ttfamily\small
Question:\par
<benchmark question>\par
\par
Image path for tools: \texttt{<opaque local image path>}.\par
Use this path only when a tool or skill requires a local image path.\par
The staged filename is an opaque temporary artifact and contains no dataset or class information.\par
\par
Return only the answer. The answer should be as specific as possible.\par
}
\end{quote}
\end{tcolorbox}

\subsection{LLM-as-Judge Prompt}
\label{app:judge-prompt}

We use the following text-only prompt for strict answer judging. The judge receives no image; it compares the model output against the verified reference answer for the given question. Line wrapping is for typesetting only.

\begin{tcolorbox}[
title=Judge Prompt,
colback=white,
colframe=black!70,
arc=5pt,
fonttitle=\bfseries\ttfamily]

\textbf{System Message}

You are an impartial evaluator for visual question answering. Your task is to judge whether a model answer is semantically correct compared with the reference answer for the given question.

Be strict about specificity. A model answer is correct only when it identifies the same target as the reference answer, or provides a strictly more specific compatible answer. If the model answer is broader, more generic, or omits required fine-grained information from the reference answer, it is not correct; use \texttt{partially\_correct} or \texttt{incorrect}, but never \texttt{correct}. For example, ``Hawker 800 series (BAe 125/Hawker 800)'' is not correct for the reference answer ``Hawker 800XP''.

Accept equivalent translations, casing differences, minor punctuation differences, and harmless formatting differences only when they preserve the same fine-grained target. For questions that explicitly require exact text matching, mark correct only when the answer text matches the reference answer exactly apart from trivial whitespace.

Return only a JSON object with these keys:
\begin{itemize}
    \item \texttt{verdict}: one of \texttt{correct}, \texttt{partially\_correct}, \texttt{incorrect}, \texttt{uncertain}
    \item \texttt{score}: a number from 0 to 1
    \item \texttt{reason}: one short sentence
\end{itemize}

\vspace{4pt}
\textbf{User Message Template}

\begin{quote}
{\ttfamily\small
\{\par
\hspace{1em}``question'': ``<benchmark question>'',\par
\hspace{1em}``reference\_answer'': ``<verified answer>'',\par
\hspace{1em}``model\_answer'': ``<parsed model answer>'',\par
\hspace{1em}``model\_raw\_content'': ``<raw model response>'',\par
\hspace{1em}``model\_rationale'': ``<parsed model rationale, if available>''\par
\}\par
}
\end{quote}
\end{tcolorbox}

\newpage
\section*{NeurIPS Paper Checklist}

\begin{enumerate}

\item {\bf Claims}
    \item[] Question: Do the main claims made in the abstract and introduction accurately reflect the paper's contributions and scope?
    \item[] Answer: \answerYes{}.
    \item[] Justification: The abstract and introduction state the benchmark scope, construction pipeline, dataset size, and main empirical findings. The claims are tied to the benchmark, experiments, and analysis sections, and are phrased as observations over the evaluated systems.

\item {\bf Limitations}
    \item[] Question: Does the paper discuss the limitations of the work performed by the authors?
    \item[] Answer: \answerYes{}.
    \item[] Justification: The conclusion includes a limitations and future work paragraph. It notes that the benchmark prioritizes annotation reliability and leakage control over exhaustive domain coverage, and that the taxonomy, evidence links, regional coverage, and process-level metrics should be refreshed or expanded.

\item {\bf Theory assumptions and proofs}
    \item[] Question: For each theoretical result, does the paper provide the full set of assumptions and a complete (and correct) proof?
    \item[] Answer: \answerNA{}.
    \item[] Justification: The paper is a benchmark and empirical evaluation study. It does not present theoretical results, theorems, or proofs.

    \item {\bf Experimental result reproducibility}
    \item[] Question: Does the paper fully disclose all the information needed to reproduce the main experimental results of the paper to the extent that it affects the main claims and/or conclusions of the paper (regardless of whether the code and data are provided or not)?
    \item[] Answer: \answerNo{}.
    \item[] Justification: We report the task definition, construction protocol, prompts, tool configurations, timeout, metric, and judge prompt. Because several evaluated systems rely on closed-source APIs and web-enabled agent execution, exact bit-for-bit reproduction can depend on external service states, so we conservatively answer \answerNo{}.

\item {\bf Open access to data and code}
    \item[] Question: Does the paper provide open access to the data and code, with sufficient instructions to faithfully reproduce the main experimental results, as described in supplemental material?
    \item[] Answer: \answerNo{}.
    \item[] Justification: We describe the benchmark, data schema, source usage, prompts, and evaluation protocol. Because the benchmark combines third-party public sources, real-life images, and web evidence links with different redistribution constraints, we conservatively answer \answerNo{} for fully open access to all data and code.

\item {\bf Experimental setting/details}
    \item[] Question: Does the paper specify all the training and test details (e.g., data splits, hyperparameters, how they were chosen, type of optimizer) necessary to understand the results?
    \item[] Answer: \answerYes{}.
    \item[] Justification: The experiments section specifies the evaluated closed-book and agent settings, shared prompts, timeout, independence constraints, and strict LLM-as-judge metric. Additional prompt templates and agent tool configurations are provided in the appendix; no model training is performed.

\item {\bf Experiment statistical significance}
    \item[] Question: Does the paper report error bars suitably and correctly defined or other appropriate information about the statistical significance of the experiments?
    \item[] Answer: \answerNo{}.
    \item[] Justification: We evaluate systems on a fixed curated benchmark and report accuracy values under a common protocol. Since the main results are reported as benchmark scores rather than repeated stochastic trials with confidence intervals, we conservatively answer \answerNo{}.

\item {\bf Experiments compute resources}
    \item[] Question: For each experiment, does the paper provide sufficient information on the computer resources (type of compute workers, memory, time of execution) needed to reproduce the experiments?
    \item[] Answer: \answerNo{}.
    \item[] Justification: We report the 240-second per-example timeout and distinguish local-model and API-based evaluations. Since compute usage partly depends on external model providers and web-agent execution, we conservatively answer \answerNo{} for complete per-experiment compute accounting.

\item {\bf Code of ethics}
    \item[] Question: Does the research conducted in the paper conform, in every respect, with the NeurIPS Code of Ethics \url{https://neurips.cc/public/EthicsGuidelines}?
    \item[] Answer: \answerYes{}.
    \item[] Justification: To the best of our assessment, the work is a benchmark and evaluation study that uses public-source data and volunteer-collected real-life images with privacy redaction. No new deployed model or intervention on users is introduced.

\item {\bf Broader impacts}
    \item[] Question: Does the paper discuss both potential positive societal impacts and negative societal impacts of the work performed?
    \item[] Answer: \answerNo{}.
    \item[] Justification: We discuss the scientific motivation, benchmark limitations, and privacy redaction for real-life data. Since the paper focuses on benchmark construction and evaluation rather than deployment, broader impacts are discussed only indirectly, so we conservatively answer \answerNo{}.

\item {\bf Safeguards}
    \item[] Question: Does the paper describe safeguards that have been put in place for responsible release of data or models that have a high risk for misuse (e.g., pre-trained language models, image generators, or scraped datasets)?
    \item[] Answer: \answerYes{}.
    \item[] Justification: The benchmark construction includes leakage inspection, evidence review, filtering of uncertain samples, and privacy redaction for real-life images. The paper releases no new high-risk generative or pretrained model.

\item {\bf Licenses for existing assets}
    \item[] Question: Are the creators or original owners of assets (e.g., code, data, models), used in the paper, properly credited and are the license and terms of use explicitly mentioned and properly respected?
    \item[] Answer: \answerNo{}.
    \item[] Justification: We cite the source datasets and model families used in benchmark construction and evaluation. Because the evidence-grounded benchmark also references heterogeneous web sources and tools with source-specific terms, we conservatively answer \answerNo{} for fully explicit licensing and terms coverage.

\item {\bf New assets}
    \item[] Question: Are new assets introduced in the paper well documented and is the documentation provided alongside the assets?
    \item[] Answer: \answerYes{}.
    \item[] Justification: The paper introduces \benchmark{} and documents its problem definition, taxonomy, construction pipeline, source usage, data schema, prompts, and evaluation protocol. The benchmark records include evidence fields and source-split metadata intended to support auditing.

\item {\bf Crowdsourcing and research with human subjects}
    \item[] Question: For crowdsourcing experiments and research with human subjects, does the paper include the full text of instructions given to participants and screenshots, if applicable, as well as details about compensation (if any)?
    \item[] Answer: \answerNo{}.
    \item[] Justification: We state that real-life images were collected from volunteers and that human annotators performed label and evidence review. Since this was a dataset curation process rather than a standalone crowdsourcing experiment with a reported interface study, we conservatively answer \answerNo{}.

\item {\bf Institutional review board (IRB) approvals or equivalent for research with human subjects}
    \item[] Question: Does the paper describe potential risks incurred by study participants, whether such risks were disclosed to the subjects, and whether Institutional Review Board (IRB) approvals (or an equivalent approval/review based on the requirements of your country or institution) were obtained?
    \item[] Answer: \answerNo{}.
    \item[] Justification: The real-life images were voluntarily contributed and privacy-redacted as part of dataset curation rather than a behavioral study. Because formal review requirements depend on the applicable institutional policy, we conservatively answer \answerNo{}.

\item {\bf Declaration of LLM usage}
    \item[] Question: Does the paper describe the usage of LLMs if it is an important, original, or non-standard component of the core methods in this research? Note that if the LLM is used only for writing, editing, or formatting purposes and does \emph{not} impact the core methodology, scientific rigor, or originality of the research, declaration is not required.

    \item[] Answer: \answerYes{}.
    \item[] Justification: LMMs are central to both benchmark construction and evaluation: they are used for model-hard filtering, closed-book and agent evaluation, and strict LLM-as-judge scoring. The relevant model usage, prompts, and judge protocol are described in the benchmark, experiments, and appendix sections.

\end{enumerate}

\end{document}